\documentclass{article}
\usepackage[margin=1in]{geometry}
\usepackage[utf8]{inputenc}
\usepackage[T1]{fontenc}
\usepackage{courier}
\usepackage{fvextra}
\usepackage{amssymb,amsmath}
\usepackage{array,tabularx,booktabs}
\usepackage{microtype}
\usepackage{xcolor}
\usepackage{hyperref}
\usepackage{url}
\hypersetup{
  breaklinks=true,
  pdfauthor={Yunqi Lu, Tyler Baumgartner, Nikhil Johri, Brandon Tai, Candice Fan, Luc Debaupte, Ruben Aguilar, Bill Wang, Yi Zhong},
  pdftitle={Continue, Adapt, or Yield: In-Turn Adaptation to Overlapping Speech in Full-Duplex Agents},
  colorlinks=true,citecolor=blue,urlcolor=blue,linkcolor=magenta,pdfborder={0 0 0}}
\newcolumntype{L}{>{\raggedright\arraybackslash}X}
\newcommand{\source}{\textsc{Source}}
\newcommand{\personaplex}{\textsc{PersonaPlex}}
\title{Continue, Adapt, or Yield: In-Turn Adaptation to\\Overlapping Speech in Full-Duplex Agents}
\author{\textbf{Yunqi Lu, Tyler Baumgartner, Nikhil Johri,}\\
\textbf{Brandon Tai, Candice Fan, Luc Debaupte,}\\
\textbf{Ruben Aguilar, Bill Wang, and Yi Zhong}\\
Besimple AI, San Mateo, CA\\
\texttt{yi@besimple.ai}}
\date{}
\begin{document}
\maketitle

\begin{abstract}
Full-duplex evaluation often emphasizes whether an agent keeps speaking or stops. That binary cannot express a third response humans use routinely: continuing to speak while incorporating what the listener just contributed. The contribution may be a missing word, a correction or a clarification. We introduce Duplex Cue, an evaluation of this \emph{in-turn adaptation} in full-duplex voice agents. Duplex Cue separates listener intent (backchannel, collaboration, or interruption) from speaker behavior: continuing unchanged, adapting within the turn, or yielding. Adaptation includes acknowledgment as well as content revision. In a single-model case study using 300 human-confirmed cues from unscripted English conversations, we compare recorded human responses with PersonaPlex continuations generated while replaying the listener's audio. We retain 208 pairs with the ongoing speaker active at cue onset and a scorable response in each condition. On the 66 collaborative pairs, recorded speakers adapt in 68.2\% of cases, compared with 34.8\% for PersonaPlex. The model otherwise continues unchanged (42.4\%) or yields (22.7\%). These findings show why evaluating natural voice interaction requires measuring how an agent responds to a listener's contribution as well as whether it keeps speaking.
\end{abstract}

\section{Introduction}
\label{sec:introduction}
In one of our recorded conversations, a speaker searching for a name trails off mid-word; the listener supplies its first syllable, then the full name, while the speaker is still mid-thought. The speaker completes that name and continues discussing it, treating the contribution as something to take up rather than a request to stop. This contribution helps shape the ongoing utterance without requiring a new turn. A full-duplex agent facing the same cue could continue its original train of thought, incorporate the suggestion, or stop to let the listener speak. Each response implies a different interaction.

We study \emph{in-turn adaptation}: an evidenced response to an overlapping contribution within an ongoing turn, including acknowledgment, rewording, and substantive revision. This distinction matters for full-duplex voice agents that must listen while speaking. Stopping promptly can be useful when a listener claims the floor, but a collaborative completion or clarification may instead invite the agent to adjust its current utterance. Conversely, uninterrupted speech can conceal a missed contribution. 

Existing frameworks do not cleanly isolate adapting while retaining the floor. A recent full-duplex survey~\cite{lu2026duplex} illustrates this omission structurally. It defines Repair on its intent axis, but Repair appears in none of its diagnostic cells, state-machine transitions, or coverage tables. Its response axis (Continue, Stop, Wait, Backchannel, Ignore, Initiate) has no value for continuing with revision.

Duplex Cue describes each event with two separate judgments. In two-person conversation, assuming Speaker A has the floor, Speaker B's contribution is labeled Backchannel, Collaboration, or Interruption according to its local intent. Speaker A's response is labeled Continued, Adapted, or Yielded. The resulting matrix distinguishes the contribution made from the behavior it receives. Recorded responses provide a reference for how people handled the same cues; differences from that reference are interpreted through the individual response categories.

We make three contributions: an operational distinction between in-turn uptake and a handoff; an evidence-linked inventory of 2,591 cues from 20.32 hours of unscripted conversation; and a paired PersonaPlex case study showing how unchanged continuation and yielding each contribute to a large adaptation gap on collaborative cues. The evaluation uses conditional dialogue continuation with a recorded partner~\cite{ntpp,turnguide}. Its primary output is the response distribution for each cue type.

\section{Related work}
\label{sec:related}
\paragraph{Listener contributions and overlapping talk.}
Listeners help shape an utterance while it is being produced. Goodwin distinguishes continuers from assessments within extended turns~\cite{goodwin1986}, while Lerner examines collaborative completions and how the original speaker accepts, incorporates, or disregards them~\cite{lerner2004}. Jefferson distinguishes exposed correction from correction embedded in ongoing talk~\cite{jefferson1987}. Overlap itself does not establish competition for the floor: Schegloff describes both noncompetitive overlap and the practices participants use to resolve competing talk~\cite{schegloff2000}. Bavelas et al. also provide experimental evidence that distracting listeners reduces their content-specific responses and impairs narrators' storytelling~\cite{bavelas2000}. These findings motivate examining both a listener's contribution and the speaker's response. Conversation-analytic overlap taxonomies and recent full-duplex work further distinguish the timing and function of overlapping speech~\cite{gervits2018,lu2026duplex}. Duplex Cue draws on these distinctions to define operational cue-intent and response labels for voice-agent evaluation.

\paragraph{Full-duplex and turn-taking evaluation.}
Existing benchmarks measure several aspects of responding to overlapping speech. Full-Duplex-Bench evaluates turn behavior and response semantics~\cite{fdb1,fdb15}, with later releases testing interactive correction handling~\cite{fdb2} and tool use under disfluency~\cite{fdb3}. HumDial-FDBench evaluates interruption and rejection in performed dialogues~\cite{humdial}, while EchoChain tests state updates under interruptions~\cite{echochain}. FD-Bench and MTR-DuplexBench assess response quality across multi-round interactions~\cite{fdbench,mtr}; FLEXI includes semantic topic shifts and agent-initiated interruptions~\cite{flexi}. Related work evaluates semantic interruption points and stopping costs~\cite{sid}, turn-event understanding and prediction~\cite{talkingturns}, and event taxonomies for end-of-turn and interruption detection~\cite{turnbench}. Duplex Cue focuses on whether the ongoing speaker continues unchanged, adapts within the turn, or yields the floor. It measures these responses separately for backchannels, collaborative contributions, and interruptions from unscripted conversation, distinguishing uptake within the current turn from uptake after a handoff.

\paragraph{Conditional dialogue continuation.}
dGSLM and SyncLLM generate continuations from real two-channel dialogue prefixes~\cite{dgslm,syncllm}. NTPP and TurnGuide evaluate a setting in which one speaker's recorded audio continues while the model generates the other's response~\cite{ntpp,turnguide}. We adopt this setting, selecting generation cutovers immediately before listener contributions and comparing generated responses with the corresponding recorded responses. Our case study evaluates PersonaPlex, which adds voice and role conditioning to the Moshi speech-text architecture~\cite{moshi,personaplex}.

\section{Definitions}
\label{sec:space}
In a two-person conversation, Speaker A holds the floor, and Speaker B contributes while A is still mid-thought. We refer to B's contribution as the \emph{cue} and A's subsequent behavior as the \emph{response}. Each event is described by two separate labels: B's cue intent and A's observed response.

\subsection{Cue intent}
\label{sec:cue-intent}
A \emph{Backchannel} supports A's ongoing speech without calling for a change in its content or direction. A \emph{Collaboration} offers an answer, correction, clarification request, constraint, or completion to help shape A's ongoing utterance without claiming the floor. An \emph{Interruption} attempts to take or retain the floor from A.

\subsection{Observed response}
\label{sec:observed-response}
\emph{Continued} describes A proceeding as if B had not contributed. \emph{Adapted} describes A responding to B within the ongoing turn through acknowledgment, rewording, or substantive revision. Adaptation concerns local uptake rather than the correctness of a correction. \emph{Yielded} describes A handing over the floor in response to B.

\section{Methods}
\label{sec:evaluation}
\subsection{Corpus and cue review}
The corpus contains 80 dual-channel English conversations totaling 20.32 hours, with 39 participants aged 18--64. Recordings average 15.24 minutes (range 14.99--30.36). Collection instructions requested natural, unscripted conversation.

Energy-based voice activity detection (VAD) identified intervals on each track. ElevenLabs Scribe v2 supplied transcripts and word timestamps. Words were assigned to the VAD interval containing their midpoint or the nearest interval, with ties assigned earlier; timestamps were clipped or shifted to fit that interval. Cross-speaker segment comparisons proposed contributions beginning while the earlier speaker was active or less than 0.5 seconds after their last word.

Individual model review screened all 10,478 proposals for eligibility. A had to hold the floor and be mid-thought when B began, with evidence that A's turn continued into B's first word. Pauses of less than 0.5 seconds between A's words were allowed, including cues that began during a brief pause within A's turn. A also had to be mid-sentence at the generation cutover, 80 ms before B's onset. These criteria applied to the recorded conversation; generated A activity was assessed separately.

Eligible contributions were then classified by intent, with ambiguous, unusable, duplicate, and out-of-taxonomy cases rejected. Reviewers considered B's developing contribution in the surrounding source speech, including A's subsequent response, and determined B's intent independently of A's compliance. Each decision retained timed evidence, a rationale, and the closest alternative. For interruptions, the reviewer also identified the earliest B word making the floor claim clear. Generation still began relative to B's first word. Recovered cue-review sessions used OpenAI Codex with \texttt{gpt-6-astra} at medium or high reasoning. The inventory contained 1,896 Backchannels, 361 Collaborations, and 334 Interruptions; Appendix~\ref{sec:review-details} reports the rejection counts.

A human reviewer labeled a balanced audio sample of 60 cues, with the model labels hidden. Agreement was 49/60 (81.7\%): 19/20 Backchannels, 14/20 Collaborations, and 16/20 Interruptions. Seven of eleven disagreements crossed the Collaboration--Interruption boundary. This validation sample came from the first ten conversations and was kept separate from the evaluation-selection queue (Appendix~\ref{sec:human-cue-validation}).

\subsection{Selection and eligibility}
\label{sec:cue-selection}
Multiple human reviewers conducted an additional audio review of a broader pool of candidate cues, with model-assigned labels hidden. Review prioritized clear temporal overlap with A's ongoing speech and fewer additional contributions from B beyond the target cue, reducing ambiguity in attributing A's response to that cue. The resulting cohort contained 300 human-confirmed trials, balanced across the three cue categories, from 65 conversations and 38 participants.

B's recorded channel remained audible throughout the generation window. Response judgments were anchored to the selected cue. For the primary comparison, we further required A to be active at cue onset in both conditions and both responses to be scorable. Table~\ref{tab:selection} accounts for the resulting sample. For each cue category, we compare the counts and proportions of Continued, Adapted, and Yielded across recorded and generated responses to the same cues. Selection therefore defines a controlled set of human-confirmed cues rather than preserving their prevalence in the full corpus.

\begin{table}[t]
\centering\small
\begin{tabular}{lrrrr}
\toprule
Stage & Backchannel & Collaboration & Interruption & Total \\
\midrule
Selected paired trials & 100 & 100 & 100 & 300 \\
Activity-ineligible pairs & 16 & 33 & 38 & 87 \\
Both responses active & 84 & 67 & 62 & 213 \\
Unusable response among active pairs & 3 & 1 & 1 & 5 \\
Primary paired trials & 81 & 66 & 61 & 208 \\
\bottomrule
\end{tabular}
\caption{Eligibility accounting for the 300 selected trials. Activity and transcript eligibility after generation yield 208 paired trials (416 responses) for the primary comparison.}
\label{tab:selection}
\end{table}

\subsection{Conditional generation}
\label{sec:conditional-continuation}
We converted A's full source track to ElevenLabs' fixed premade George voice before model conditioning, honoring the contributors' no-cloning commitment without constructing a participant-specific voice model. Converted tracks were reused across cutovers. B's input and A's recorded reference response retained their original voices.

We follow the conditional dialogue-continuation approach used in NTPP and TurnGuide~\cite{ntpp,turnguide}: the model continues one speaker's channel while the other speaker's recorded audio remains the input. For each trial, we conditioned \texttt{nvidia/personaplex-7b-v1} on up to 120 seconds of synchronized audio from both speakers immediately preceding the generation cutover. B's recorded audio was supplied to the model's input stream, while the audio tokens from A's converted speech were supplied to its output stream through teacher forcing. This established the preceding conversation as the model's listening and speaking history, with A occupying the role the model would continue. At the cutover, 80 ms before B's cue onset and clipped at zero, we stopped supplying A's recorded tokens and allowed the model to generate its own continuation on the same output stream for ten seconds. B's audio, including the target cue, continued on its original timeline throughout generation. No cue label, task category, or persona text was supplied. A's original recorded continuation over the same window formed the \source{} condition. Appendix~\ref{sec:reproducibility} specifies conversion and inference settings.

\subsection{Response review and human validation}
\label{sec:response-review}
We used ElevenLabs Scribe v2 to obtain word-timed transcripts of the evaluated output. Each review packet combined up to 15 seconds of preceding input and output transcripts with the ten-second output to be scored and B's input over that same interval. This provided the scorer with A's preceding conversational trajectory, B's developing contribution, and A's response on a shared timeline. Model-response packets contained only the model's continuation after the trial start; the source continuation was withheld.

Each packet was reviewed individually by OpenAI Codex with \texttt{gpt-6-astra} at medium reasoning using a response-classification skill: a written set of instructions specifying the label definitions, evidence requirements, and structured review format. The same skill was applied to both conditions, with the cue label, expected action, and paired response judgment hidden. It assigned Continued when A proceeded without observable uptake of B's cue, Adapted when A acknowledged or incorporated B's contribution within the ongoing turn, and Yielded when A relinquished the floor in response to B. An established handoff took precedence over any subsequent uptake. The skill required a rationale grounded in timed transcript evidence, consideration of the closest alternative label, and a confidence judgment. Responses with no speech or insufficient usable evidence were excluded and reported separately. Appendix~\ref{sec:review-details} provides the full decision rules.

A human reviewer manually reviewed 150 responses, with targeted adjudication of prior disagreements. Using the latest adjudicated ratings, scorer--human agreement was 108/142 (76.1\%; 95\% CI: 68.4--82.3\%) among comparable responses, with Cohen's $\kappa=0.641$. Six human \texttt{not\_overlap} judgments and two additional responses excluded as unusable by the scorer were omitted; retaining all 150 ratings in the denominator gives 108/150 (72.0\%). Appendix~\ref{sec:human-response-validation} reports agreement by scorer label.

\section{Results}
\label{sec:results}

\subsection{Recorded speakers distribute across all three responses}
\label{subsec:source}

No cue type mapped cleanly onto a single response in the recorded condition
(Table~\ref{tab:matrix}). Backchannels came closest, with continuation the majority
response at 71.6\%, but even there recorded speakers adapted on 23.5\% of cues.
Collaborations drew adaptation on 68.2\% and continuation on 22.7\%. Interruptions
were the least concentrated of the three.

The interruption result bears directly on how overlap handling is scored. Recorded
speakers yielded on only 24 of 61 interruption cues (39.3\%), and adapted within the
ongoing turn on 28 (45.9\%). The most common human response to an evident floor claim
was therefore not to hand over the floor, but to answer or acknowledge while continuing
to speak. Existing benchmarks treat prompt yielding as the target behavior for
interruptions and measure stop latency or overtalk against it; under that criterion,
the majority of recorded human responses here would count as failures.

Two qualifications apply. Our interruption class is defined by B's evident attempt to
take or retain the floor, whether or not A complied, so it includes attempts that did
not succeed; a class restricted to completed floor transfers would show a higher yield
rate by construction. And the ten-second generation window may truncate handoffs that
resolve later, which would understate yielding in both conditions.

These distributions are the reference against which we read the model condition. They
also motivate reporting the full matrix rather than a single mapping score: a scoring
rule that treats one response as correct per cue type would misclassify a large share
of recorded human behavior.

\begin{table}[t]
\centering\small
\begin{tabular}{llrrrr}
\toprule
Condition & Cue type & Continued & Adapted & Yielded & $N$ \\
\midrule
\source & Backchannel & 58 (71.6\%) & 19 (23.5\%) & 4 (4.9\%) & 81 \\
 & Collaboration & 15 (22.7\%) & 45 (68.2\%) & 6 (9.1\%) & 66 \\
 & Interruption & 9 (14.8\%) & 28 (45.9\%) & 24 (39.3\%) & 61 \\
\midrule
\personaplex & Backchannel & 57 (70.4\%) & 17 (21.0\%) & 7 (8.6\%) & 81 \\
 & Collaboration & 28 (42.4\%) & 23 (34.8\%) & 15 (22.7\%) & 66 \\
 & Interruption & 10 (16.4\%) & 20 (32.8\%) & 31 (50.8\%) & 61 \\
\bottomrule
\end{tabular}
\caption{Observed response distributions on identical active, scorable pairs. Cells give counts and row percentages. Adapted includes responsive acknowledgment as well as content revision.}
\label{tab:matrix}
\end{table}

\subsection{PersonaPlex adapts less often on collaborative cues}
\label{subsec:model}

On the 66 active, scorable Collaboration pairs, recorded speakers adapted in 45 cases
(68.2\%) and PersonaPlex in 23 (34.8\%), a difference of 33.3 percentage points.
PersonaPlex continued unchanged in 28 cases (42.4\%) and yielded in 15 (22.7\%); the
corresponding source counts were fifteen and six. The model's lower adaptation rate
therefore reflects both unchanged continuation and handoff behavior. An aggregate
non-yield rate would combine adaptation with unchanged continuation and obscure this
distinction.

The paired breakdown supports the descriptive comparison: 20 of 23 model adaptations
occurred on trials where the source also adapted; 25 pairs had source adaptation alone,
and 18 had adaptation in neither condition.

Table~\ref{tab:examples} illustrates how the response labels apply across four
Collaboration trials, chosen to span the observed response combinations rather than
to showcase only failure or only success. In trial 0135, the recorded speaker takes up
B's suggested name; PersonaPlex continues with a different guess. In trial 0124, the
recorded speaker answers B's confirmation question before B finishes, then continues
speaking; PersonaPlex instead abandons an unfinished phrase and resumes only after the
question ends. In trial 0175, both speakers adapt but develop the contribution
differently, illustrating that Adapted does not imply agreement on content. Trial 0191
reverses the more common pattern: the recorded speaker yields the floor to B's
clarification question, while PersonaPlex answers it directly before B finishes
asking.

\begin{table}[t]
\centering\small
\begin{tabularx}{\columnwidth}{p{.2\columnwidth}LL}
\toprule
Collaborative contribution & Recorded response & PersonaPlex response \\
\midrule
Name completion: B supplies a name's initial syllable, then the name. & A completes that name at $+0.10$--$0.68$ s and continues discussing its pronunciation. \textbf{Adapted.} & A continues a different candidate name at $0.00$--$1.36$ s without taking up B's suggestion. \textbf{Continued.} \\
\midrule
Scheduling check: B asks whether the proposed stopping time works ($0.00$--$1.36$ s). & A confirms at $+0.92$--$1.98$ s, beginning before B finishes, then continues. \textbf{Adapted.} & A cuts off an unfinished phrase at $+0.06$ s and resumes at $+2.14$ s, after B finishes. \textbf{Yielded.} \\
\midrule
Assembly description: B characterizes a task as involving a few simple parts. & A breaks off, gives a brief acknowledgment at $+1.9$--$2.1$ s, then resumes its own account of calling a friend for help. \textbf{Adapted.} & A acknowledges B's specific phrase at $+2.0$--$4.3$ s, states it does not have the needed parts on hand, and proposes a store trip to get them at $+5.8$--$7.9$ s. \textbf{Adapted.} \\
\midrule
Field clarification: B asks which specific date field A means. & A stops mid-word at cue onset, stays silent through B's roughly three-second question, and answers only after B finishes, at $+3.6$ s. \textbf{Yielded.} & A finishes its own clause at $+0.9$ s, then directly corrects B's proposed reading while B is still asking, at $+3.4$--$8.8$ s. \textbf{Adapted.} \\
\bottomrule
\end{tabularx}
\caption{De-identified, timed paraphrases of four evaluated Collaboration trials (0135, 0124, 0175, 0191), chosen to span the observed response combinations. Timings are relative to B's cue onset. These illustrate the recorded model-judge labels; they are not verbatim transcripts or additional human validation.}
\label{tab:examples}
\end{table}

On the other two cue types, the conditions diverged in opposite directions.
Backchannels produced near-identical distributions (continuation 71.6\% source,
70.4\% model). On interruptions, PersonaPlex yielded more often than the source
(50.8\% versus 39.3\%) and adapted less (32.8\% versus 45.9\%). This is the same
asymmetry seen on collaborations, in a category where prompt yielding is more often
appropriate.

Response agreement was lowest on the \textsc{Adapted} label (68.6\%,
Table~\ref{tab:human-response}) and cue agreement lowest on Collaboration
(14/20, Appendix~\ref{sec:human-cue-validation}), with seven of eleven cue disagreements
crossing the Collaboration--Interruption boundary. The comparison in this subsection
therefore rests on the least reliable cell of both label sets.

\subsection{Coverage}
\label{subsec:coverage}

Table~\ref{tab:coverage} reports activity and scorability across all 300 trials. PersonaPlex had 86 inactive-at-cue judgments and 46 unusable responses among 300 reviewed outputs. The primary comparison excludes those cases through the shared pair criteria, so the 34.8\% adaptation rate is conditional on both activity and scorability. Appendix~\ref{sec:extended-results} gives category-specific coverage.

\begin{table}[t]
\centering\small
\begin{tabular}{lrr}
\toprule
Diagnostic & \source & \personaplex \\
\midrule
Reviewed responses & 300 & 300 \\
Unusable responses & 4 & 46 \\
Empty ASR / available measurements & -- & 35/300 \\
Empty VAD / available measurements & -- & 7/300 \\
Judged inactive at cue onset & 3 & 86 \\
Low-confidence reviews & 3 & 18 \\
\bottomrule
\end{tabular}
\caption{Coverage over all reviewed responses. Categories overlap. ASR/VAD measurements are available for zero source responses and all model responses; missing measurements are omitted from those denominators.}
\label{tab:coverage}
\end{table}

\section{Discussion and limitations}
\label{sec:limitations}
Duplex Cue makes a practical interaction distinction measurable: retaining the floor while responding to a listener differs from retaining it while proceeding unchanged. The Collaboration results show that yielding is also common in the evaluated model condition. This suggests evaluating content uptake and floor behavior together when developing full-duplex agents. Semantic task scoring can further distinguish acknowledgment from successful incorporation within the Adapted category.

\paragraph{What existing metrics would score.}
Consider trial 0124 (Table~\ref{tab:examples}): B asks whether a proposed stopping time works before finishing the question; the recorded speaker confirms while B is still talking, then continues. Full-Duplex-Bench's turn-overlap statistics~\cite{fdb1} would register this as a takeover (B began speaking during A's turn) without distinguishing whether A went on to respond to the content or simply ceded the floor. HumDial-FDBench's scheme~\cite{humdial} classifies a mid-utterance correction of this kind as an Interruption and scores only Respond as correct; under that criterion, PersonaPlex's actual behavior on this trial (an unfinished phrase followed by silence until B's question ends) is scored as a failure to respond, with no credit for the utterance that follows the handoff. Full-Duplex-Bench v1.5's Respond/Resume categories~\cite{fdb15} have no cell for a response that answers a question and then continues the prior utterance, which is what the recorded speaker did; the closest available label would either drop the confirmation or misdescribe the continuation as an unrelated resumption. Each existing scheme therefore either collapses this distinction or scores the recorded human response as an error.

\paragraph{Annotation and validation.}
Cue intent was not labeled blind to Speaker A's subsequent response: reviewers could see A's behavior when judging whether B's contribution was a Collaboration or an Interruption. Collaboration is partly identified by the source going on to adapt, so some portion of the 68.2\% adaptation rate on Collaboration cues is definitional rather than an independent behavioral finding. A blind relabeling that withholds post-cue audio would separate cue intent from response outcome and is the cheapest check against this threat. Further validation can also examine how consistently the response categories transfer across annotators and evidence formats more broadly. The present study reports agreement with one human reviewer; additional independent reviewers would allow direct measurement of human--human agreement. Systematic audio review alongside transcripts, including currently excluded responses, would help characterize sensitivity to transcription quality and clarify floor boundaries.

\paragraph{Controlled and interactive evaluation.}
The current comparison evaluates recorded speakers with their original conversational context alongside model continuations conditioned on bounded, voice-converted history. Future experiments can vary history length and voice conversion to separate their contributions to response behavior. Causal voice conversion would also remove access to future source audio during preprocessing, while timing and prosody checks could quantify changes introduced by conversion. No-cue controls would help distinguish cue-driven adaptation from spontaneous continuation. Interactive partners would extend the evaluation beyond fixed listener timing, and longer response windows would capture later uptake following a handoff.

\paragraph{Broader evaluation.}
Applying Duplex Cue to additional models, repeated generations, languages, and independently sampled conversations would test how broadly the observed response patterns hold. The present results cover one checkpoint and one generation per trial in English, with repeated participants. The primary comparison uses human-confirmed cues and pairs that meet activity and scorability criteria; it describes behavior within that subset. Broader sampling with explicit accounting for generation failures would connect these conditional results to overall system performance and natural cue frequencies. Replication on openly available or independently licensed recordings would broaden access beyond the current corpus, which requires authorized artifacts and external providers.

\section{Conclusion}
Duplex Cue makes in-turn adaptation an explicit target for full-duplex evaluation: how a speaker responds to a listener's contribution while retaining the floor. By separating listener intent from observed response, it distinguishes unchanged continuation, adaptation within the turn, and yielding across cues from unscripted conversation. On 66 active, scorable collaborative pairs, recorded speakers adapted in 68.2\% of cases, compared with 34.8\% for PersonaPlex. The model's remaining responses divided between unchanged continuation and yielding, revealing distinct patterns that an aggregate stop/continue measure would obscure.

This single-model study with a recorded partner demonstrates the diagnostic value of measuring uptake and floor behavior together. Duplex Cue provides a framework for examining that distinction across models and interaction settings. Evaluating full-duplex conversation should therefore ask how a listener's contribution shapes the utterance already underway, alongside whether the speaker continues or yields.

\section*{Ethics statement}
We reused privately licensed recordings without new recruitment. An audit verified pre-capture acceptance of research and processing terms for all 39 participants and 160 participant--recording pairs, along with collection instructions and credit awards, though this retrospective evaluation of previously collected recordings did not undergo a formal ethics-board review or exemption determination. Contributors were promised no voice cloning; fixed stock-voice conversion of A supplied model history without constructing participant-specific voice models.

ElevenLabs processed original A tracks for conversion and evaluated audio for transcription; PersonaPlex on Modal received converted A history and recorded B; Codex reviewed transcripts. Audio, participant transcripts, reviews, and consent evidence remain access-controlled. The examples are de-identified paraphrases. Voice conversion does not guarantee anonymity; voice impersonation, deepfake, and identification from conversational content remain risks. Codex assisted annotation, response judgments, and manuscript preparation. The authors take responsibility for the evidence and claims.

\clearpage\appendix
\section{Annotation and response decision rules}
\label{sec:review-details}
These rules define the individual model judgments of cue eligibility, intent, and response. The workflow prompts are retained as Markdown artifacts in the repository.

\subsection{Cue eligibility and intent}
Reviewers examined both speakers' timed words and surrounding syntax. A proposed event was eligible only when A had an established floor, maintained continuity into B's onset under the less-than-0.5-second gap tolerance, was mid-thought at that onset, and was mid-sentence at the cutover 80 ms earlier. False or unresolved eligibility conditions were rejected. Equal onsets, natural turn transitions, apparent overlaps caused by segment padding, and earlier speakers who were themselves backchanneling did not qualify. The cutover was not shifted to rescue an event.

For eligible events, reviewers assessed B's full developing contribution. Backchannels supported A's trajectory without a material change or floor claim. Collaborations supplied an answer, correction, clarification, constraint, or completion without a floor claim. Interruptions attempted to take or retain A's floor, whether or not A complied. A contribution beginning as an acknowledgment could become an interruption. If competing intent readings remained plausible, the event was rejected. Repeated views of one contribution were rejected as duplicates. Accepted judgments cited A and B words and explained the closest alternative. Interruptions additionally identified an actual timed B token as the earliest clear floor claim. Existing candidate labels, response scores, and human answers were withheld from model cue review.

\begin{table}[h]
\centering\small
\begin{tabular}{lr}
\toprule
Primary rejection reason & Proposals \\
\midrule
Natural turn transition & 2,375 \\
No real overlap & 95 \\
No established prior floor & 3,055 \\
A not clearly mid-thought & 1,403 \\
Ambiguous cue intent & 581 \\
Intent outside taxonomy & 30 \\
Unusable transcript & 198 \\
Duplicate contribution & 150 \\
\midrule
Total & 7,887 \\
\bottomrule
\end{tabular}
\caption{Primary model-judged rejection reasons over the complete 10,478-proposal inventory. The remaining 2,591 events were accepted.}
\label{tab:rejections}
\end{table}

\subsection{Response review}
Each packet replaced post-cutover source A with the evaluated A transcript and retained source B. Reviewers first established A's pre-cue trajectory and activity at onset, then classified the response with the cue label hidden. Continued required usable evidence of proceeding as if B had not contributed. Adapted required cited post-cue A words and an explicit link to B's contribution without a prior completed handoff. Acknowledgment and minor rewording qualified; shared topic or coincidental wording did not. Yielded required evidence of relinquishing an active floor in response to B. Later silence or a later adapted reply did not change an established Yielded judgment.

Every scorable response cited timed A words in the evaluated window; Continued and Adapted required post-cue words. The reviewer explained the behavior, its relationship to B, and the closest alternative, and recorded confidence and limitations. Low confidence alone did not exclude an otherwise scorable case. The silent exclusion meant no response speech; unusable meant the evidence could not support a behavior judgment. Empty ASR or VAD output did not establish silence. Listening could distinguish those exclusion reasons, but intelligible audio without usable timed A words remained unusable for the transcript-based scorer. No excluded response received a behavior label or mapping agreement.

The VAD diagnostic used 30-ms frames, a 10-ms hop, a 30th-percentile energy threshold floored at $-55$ dBFS, a 100-ms minimum speech duration, and 50-ms padding. Alignment to VAD was available for inspection; response review retained raw word times. Floor ownership was determined from the conversational evidence rather than a VAD interval alone.

\section{Human validation}
\subsection{Cue validation}
\label{sec:human-cue-validation}
The 60-cue validation sample contained 20 cues per model-assigned category. Its sampling pool comprised 433 accepted candidates from conversations 001--010: 276 Backchannels, 77 Collaborations, and 80 Interruptions. Within each category, candidates were ranked by SHA-256 of the seed, sampling tag, and candidate identifier; the first 20 were selected without replacement. The seed was \texttt{7677426ab6e437e307c7c33236757476}. The resulting sample spanned nine conversations. A human reviewer reviewed all 60 audio clips with model labels hidden.

Agreement was 49/60 (81.7\%). Of 20 proposed Backchannels, 19 retained that label and one became Collaboration. Of 20 Collaborations, 14 retained that label, four became Interruption, and two became Backchannel. Of 20 Interruptions, 16 retained that label, three became Collaboration, and one received Other because B already held the floor. There were no uncertainty or unusable-data ratings. Four cue onsets had previously been reviewed by the same reviewer; excluding them gave 45/56 (80.4\%) agreement. The estimate describes the balanced validation sample, including disagreements in its denominator.

\subsection{Response validation}
\label{sec:human-response-validation}
A human reviewer completed a manual review of 150 responses, with targeted adjudication of prior disagreements. The latest adjudicated ratings include 144 human Continued, Adapted, or Yielded labels and six \texttt{not\_overlap} judgments. Two of the 144 were excluded as unusable by the current scorer, leaving 142 comparable responses. Agreement was 108/142 (76.1\%; 95\% Wilson CI: 68.4--82.3\%), with Cohen's $\kappa=0.641$. Including all 150 ratings in the denominator gives 108/150 (72.0\%). These adjudicated ratings are not a fresh independent review of the current scorer. Table~\ref{tab:human-response} gives the breakdown by current scorer label.

\begin{table}[h]
\centering\small
\begin{tabular}{lrr}
\toprule
Scorer label & Agreements / comparable responses & Agreement \\
\midrule
Continued & 44/57 & 77.2\% \\
Adapted & 35/51 & 68.6\% \\
Yielded & 29/34 & 85.3\% \\
All & 108/142 & 76.1\% \\
\bottomrule
\end{tabular}
\caption{Scorer--human response agreement in the 150-item manual review. Six human \texttt{not\_overlap} judgments and two additional scorer-unusable responses are excluded from the comparable-response denominators.}
\label{tab:human-response}
\end{table}

\section{Detailed paired results}
\label{sec:extended-results}
Mapping agreement is a secondary descriptive statistic: Backchannel maps to Continued, Collaboration to Adapted, and Interruption to Yielded. For the primary paired trials, the rate in a condition is the fraction of trials whose observed response equals the mapped action. The source rate was 127/208 (61.1\%) and PersonaPlex's was 111/208 (53.4\%). These totals combine different cue behaviors and should be read alongside the full response matrices.

\begin{table}[h]
\centering\small
\begin{tabular}{lrrrr}
\toprule
Cue type & Both agree & Source only & Model only & Neither \\
\midrule
Backchannel & 47 & 11 & 10 & 13 \\
Collaboration & 20 & 25 & 3 & 18 \\
Interruption & 14 & 10 & 17 & 20 \\
All & 81 & 46 & 30 & 51 \\
\bottomrule
\end{tabular}
\caption{Paired agreement with the cue-to-response mapping. Columns partition the 208 primary pairs and describe mapped actions, not preference judgments.}
\label{tab:paired}
\end{table}

\begin{table}[h]
\centering\small
\begin{tabular}{llrrrr}
\toprule
Condition & Cue type & Empty ASR & Empty VAD & Inactive & Low conf. \\
\midrule
\source & Backchannel & -- & -- & 0 & 0 \\
 & Collaboration & -- & -- & 0 & 2 \\
 & Interruption & -- & -- & 3 & 1 \\
\personaplex & Backchannel & 5/100 & 1/100 & 16 & 2 \\
 & Collaboration & 7/100 & 1/100 & 33 & 10 \\
 & Interruption & 23/100 & 5/100 & 37 & 6 \\
\bottomrule
\end{tabular}
\caption{Coverage across all reviewed responses by cue category. Fractions count empty outputs over available measurements; -- denotes unavailable diagnostics. Inactivity counts cover all reviews. Columns overlap.}
\label{tab:cue-coverage}
\end{table}

\section{Conversion and reproducibility}
\label{sec:reproducibility}
\label{sec:voice-conversion}
We used ElevenLabs Voice Changer model \texttt{eleven\_multilingual\_sts\_v2} with the premade George voice (\texttt{JBFqnCBsd6RMkjVDRZzb}). Each A track was converted once and reused across cutovers; the reported cohort used 98 distinct converted tracks. Original source responses and B retained their recorded voices.

Tracks were normalized to 24-kHz mono PCM16 and partitioned near quiet points approximately every 240 seconds, searching within ten seconds of each boundary and including up to one second of overlap on each side. Lossless FLAC carried segments to the provider. Outputs were trimmed to their central source-coordinate intervals and assembled into a full-length WAV. A raw duration change exceeding 80 ms rejected a segment; smaller differences were padded or trimmed at its end without time stretching. This avoided cumulative duration drift while allowing local word-timing or prosodic changes.

Conversion settings were stability 0.5, similarity boost 0.75, style 0, speaker boost enabled, and background-noise removal disabled. The seed was 42,424,242 plus segment index. The converter received the full source track. PersonaPlex received only preceding converted A context, with silence staged for A after cutover and recorded B continuing on its original timeline.

The runner used 24-kHz mono PCM and 80-ms frames on an NVIDIA A100 40GB configuration. Prompt length was rounded down to whole frames while preserving the requested cutover. Generation comprised 125 frames; two delay-flush frames were excluded from saved output. Audio/text temperatures were 0.8/0.7, top-$k$ values 250/25, and the generation seed was 42,424,242 plus the zero-based trial index. The PersonaPlex checkpoint revision was \texttt{fdaf4090a61cb315c138a1faee287ffd6c716309}.

Cue and response artifacts retain timed evidence, individual rationales and audio fingerprints. The reported analysis uses the 300-trial cohort and evidence digest \texttt{9d1575a7d056b3f66ed7977e895736ce26f73b8e4f01bf8e4ee540ee956bcafb}. Saved converted WAVs define the evaluated audio inputs because fixed provider seeds need not reproduce them bit-for-bit.

The corpus and evaluation repository are access-controlled. Authorized reproduction requires the cohort manifest, source and converted audio, inference settings, and saved review evidence, as well as access to external providers. PersonaPlex uses the NVIDIA Open Model License Agreement and its reference code is MIT-licensed. Participant recordings and transcripts are excluded from manuscript distribution.

\end{document}